\documentclass[runningheads]{llncs}
\usepackage[para]{footmisc}
\usepackage{graphicx}
\usepackage{hyperref}

\usepackage{booktabs}
\usepackage{multirow}
\usepackage{makecell}
\usepackage{xcolor}
\usepackage{pifont}
\usepackage{xspace}
\newcommand{\YES}{\ding{51}\xspace}
\newcommand{\NO}{{\color{gray!75}\ding{55}}\xspace}
\newenvironment{citemize}{\begin{list}{\textbf{--}}{\topsep=.2\smallskipamount\itemsep=0pt\parsep=0pt\labelwidth=0.75em}}{\end{list}}
\makeatletter
\renewcommand\subsubsection{\@startsection{subsubsection}{3}{\z@}%
                       {-12\p@ \@plus -4\p@ \@minus -4\p@}%
                       {-0.5em \@plus -0.22em \@minus -0.1em}%
                       {\normalfont\normalsize\bfseries\boldmath}}
\makeatother

\begin{document}
\def\doi#1{\url{https://doi.org/#1}}
\title{Practical End-to-End Optical Music Recognition for Pianoform Music}
\author{Jiří Mayer\orcidID{0000-0001-6503-3442} \and
Milan Straka\orcidID{0000-0003-3295-5576} \and
Jan~Hajič~jr.\orcidID{0000-0002-9207-567X} \and
Pavel Pecina \orcidID{0000-0002-1855-5931}}
\authorrunning{J. Mayer et al.}
\institute{Charles University, Faculty of Mathematics and Physics\\ Institute of Formal and Applied Linguistics, Prague, Czech Republic\\
\email{\{mayer,straka,hajicj,pecina\}@ufal.mff.cuni.cz}
}
\maketitle

\begin{abstract}
The majority of recent progress in Optical Music Recognition (OMR) has been achieved 
with Deep Learning methods, especially models following the end-to-end paradigm,
reading input images and producing a linear sequence of tokens. 
Unfortunately, many music scores, especially piano music, cannot be easily converted to a linear sequence.
This has led OMR researchers to use custom linearized encodings,
instead of broadly accepted structured formats for music notation.
Their diversity makes it difficult to compare the performance of OMR systems directly.
To bring recent OMR model progress closer to useful results: 
(a) We define a sequential format called Linearized MusicXML, allowing to train an end-to-end model directly
and maintaining close cohesion and compatibility with the industry-standard MusicXML format.
(b) We create a dev and test set for benchmarking typeset OMR with MusicXML ground truth based on the OpenScore Lieder corpus. 
They contain 1,438 and 1,493 pianoform systems, each with an image from IMSLP.
(c) We train and fine-tune an end-to-end model to serve as a baseline on the dataset and employ the TEDn metric to evaluate the model. 
We also test our model against the recently published synthetic pianoform dataset GrandStaff
and surpass the state-of-the-art results.
\end{abstract}

\section{Introduction}

Optical Music Recognition (OMR), is the field that investigates how to computationally read music notation 
in documents \cite{CalvoZaragoza-2020-understanding}, 
is among the many sub-fields that have seen significant progress with end-to-end approaches to
recognition \cite{shi-etal-2017-an-end-to-end,Calvo-Zaragoza2017c,Calvo-Zaragoza2018b,Calvo-Zaragoza2019,RiosVila2022}.

This is straightforward for ``monophonic'' notation, 
where the encoded music has just one voice: its output thus consists of a single sequence, and it is analogous to text.\footnote{The complexity introduced by the two-dimensional compositional nature of music notation, as opposed to most writing systems for natural languages, is no longer a significant issue for current deep learning methods.}
However, despite plenty of use cases for such monophonic OMR, a vast amount of music -- and some of the most prominent repertoire in the world -- is written for piano.

\begin{figure*}[t]
    \centering
    \includegraphics[width=.95\hsize]{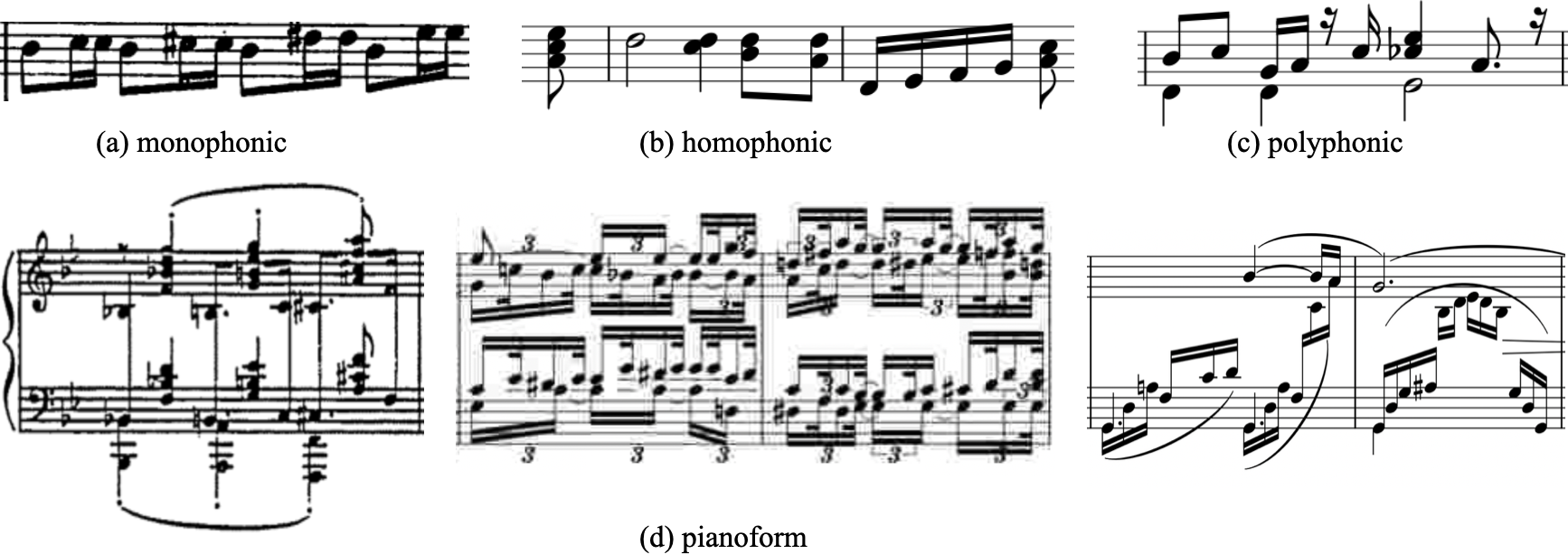}
    \caption{Typology of music notation by complexity. Monophonic scores (a) are straightforwardly encoded as sequences; for (b) homophonic scores (chords allowed, but all simultaneous notes have the same length), advance coding has been used with promising results still with CTC objective \cite{AlfaroContreras2023}. For (c) polyphony, linearization becomes necessary, and (d) pianoform music adds interaction between staffs within one grand staff and generally contains the greatest density of objects.}
    \label{fig:typology-of-notaton}
\end{figure*}

\noindent
In terms of notation complexity, pianoform music%
\footnote{The term encompasses not only piano music, but also music for organ, harpsichord, harp, vibraphone, possibly guitar, and other instruments.}
in Common Western Music Notation (CWMN) is the ``final frontier'' for OMR models \cite{CalvoZaragoza-2020-understanding} (see~Fig.~\ref{fig:typology-of-notaton}), but it too has recently seen promising results with sequence-to-sequence models \cite{Rios-Vila-etal-2023-end-to-end}.

Specifically for sequence-to-sequence models, the fact that piano music contains multiple independent voices in parallel, 
which can arbitrarily appear and disappear in a composition (even for very short segments), 
introduces an extra layer of complexity, especially on the output side. 
The problem can perhaps be compared to trying to recognize an unknown amount of texts written over each other.
Attention-based models \cite{bahdanau-etal-2015-neural}, most prominently Transformers \cite{vaswani-etal-2017-attention,Rios-Vila-etal-2023-end-to-end}, can produce an arbitrary number of outputs for a single output and do not require a monotonous alignment to exist between the input and output as in Connectionist Temporal Classification (CTC) \cite{graves-etal-2006-connectionist}, but they do require the output to be a sequence. Therefore, the ground truth for piano music must be linearized to make training and evaluation possible.
\looseness-1
However, a model that outputs such a linearized representation is by itself not particularly useful beyond
experiments. Formats that encode music notation in practice, such as MusicXML, **kern, MEI, LilyPond, or various
open or proprietary formats used to represent notation in widely used editors (MuseScore, Finale, Sibelius, Dorico, etc.), strive to capture the multi-voice structure of the encoded music -- their objective is not to be convenient for a particular
class of models -- and thus for a sequence-to-sequence OMR system to become useful in practice, its linearized
target representation must be then followed up by a de-linearization step.%
\footnote{The possible exception is **kern, which has a straightforward enough structure in text
that it can be output directly by the OMR model as plain text, and converters exist to other formats. However, it is not as widely adopted as MusicXML, and converters are imperfect.}
This step introduces further complexity: especially, 
because there is inherent randomness in the trained model's output,
it needs to be able to deal with sequences that do not necessarily lead to syntactically valid notation.

\looseness-1
OMR also differs fundamentally from OCR, in that OMR users expect not just to recover the information
about how the elements of music notation are arranged to encode a certain musical composition (what in OCR would 
be recovering the configuration of symbols on a page, termed ``reprintability'' \cite{CalvoZaragoza-2020-understanding}), but also
decode the ``musical semantics'', the composition itself -- which notes should be played at what time (termed ``replayability'', respectively \cite{CalvoZaragoza-2020-understanding}). 
All practical formats for representing music notation contain intertwined information about both these realms,
and thus a model must decode the semantics such as pitches and durations of notes 
from the configurations of the graphical elements.

This added complexity leads to another issue in turning the advances in OMR models into palpable progress: evaluation.
OMR evaluation is a difficult issue on its own \cite{Bellini2007,Byrd2015,Calvo-Zaragoza2018d,Hajicjr.2018b}.
The natural evaluation metric on sequences of tokens, Symbol Error Rate (SER), has
unclear interpretation outside of the specific encoding used by that particular system,
and does not allow for direct comparisons between different linearizations.
Because music notation is a writing system that tends to have an exception to every basic rule, 
especially in piano music \cite{Byrd2015},
it is tempting to preprocess data so that symbols and situations that appear peripheral
are left out (especially slurs and other symbols that do not directly affect how the encoded music would be
exported to MIDI), or that a priori ``easier'' datasets are assembled in the first place.
However, a transparent evaluation of a system's usefulness (as opposed to measuring just
the ability of a model to learn what is required of it) should be performed directly on the ground truth files,
also in order to show how much of the original score was discarded in this process of ``trimming down'' 
to some ``core'' subset of music notation.
While such metrics have previously been suggested, which also attempt to be more informative than SER \cite{Hajicjr.2016},
these have not seen broader adoption.

In order to design, build, and evaluate OMR systems, so that the considerable advances in the field
made in recent years thanks to end-to-end models can reach the many potential users, we tackle these challenges. 
The main contributions of this paper are therefore:\footnote{The related work for each of the contributions is discussed in their respective sections.}
\begin{citemize}
\item We propose and implement a direct linearization and de-linearization procedure for MusicXML,
the most widely adopted machine-readable music notation interchange format (Sec.~\ref{sec:lmx}),
\item collect a ``difficult'' dataset (OLiMPiC) of pianoform music notation from the OpenScore Lieder Corpus \cite{OpenScore.2018,GothamJonas2022}
with synthetic training images, but dev and test sets with real-world images from public IMSLP scans (Sec.~\ref{sec:datasets}),
\item establish an evaluation comparing MusicXML files directly with an implementation of Tree Edit Distance (TEDn),
which better correlates with the preferences of human editors \cite{Hajicjr.2016} (Sec.~\ref{sec:evaluation}), and
\item achieve state-of-the-art performance on pianoform music (Sec.~\ref{sec:results}).
\end{citemize}

\noindent While this work does present a near-complete\footnote{The only remaining step is on the input side: detect where on the page the notation is and split it into systems.} OMR system for pianoform music with state-of-the-art performance, we have little doubt that better models will soon follow. The main value of our contributions is significant improvements of OMR infrastructure that, taken together, bring considerable progress in the field much closer to application.

\section{Linearized MusicXML Encoding}
\label{sec:lmx}

In an ideal world, a recognition model will output a well-known standardized format.
Out of the available formats, we consider MusicXML to be the most practical choice 
for machine-readable music notation today, as it retains broad support among all popular notation editors 
and further tooling (such as the Music21 library), 
and while the successor MNX format\footnote{\url{https://github.com/w3c/mnx}} is being developed 
in the W3C Music Notation Community Group, no obsolescence for MusicXML is planned.
Importantly, MusicXML is also the preferred interoperability format 
for the MuseScore open-source notation editor.\footnote{\url{https://musescore.org/en/node/82366\#comment-363536}} 
Thanks to this broad support, nearly all of the music stored in a computer-readable format can be expected to have a way of being exported to MusicXML with relatively little information lost due to priority support for MusicXML conversion. 
We therefore view the ability to use MusicXML files for training OMR systems, and producing results in MusicXML, 
as a major step towards shortening the journey from improved OMR models to improved results for users.

However, while it is technically possible to train a model to output MusicXML files directly,
this is not an optimal choice. XML-based formats are tree-based and often excessively verbose. 
They often contain lots of additional information that cannot be leveraged for training: 
for example, unique element IDs, other metadata, or pixel-perfect formatting of the score. 
While it would technically be possible to train a model to output MusicXML strings directly despite these disadvantages,
MusicXML (as do all such structured formats) has strict rules on syntax and validity.
Thus, a single recognition mistake on a page can make the output document completely invalid, and not even processable
by GUI tools for post-correction that assume a valid MusicXML file on input.
One would therefore anyway have to implement some post-processing step 
that would handle the inevitable errors against XML syntax or MusicXML specification. 
Since a postprocessing step is thus necessary anyway (and in practice preprocessing as well, 
at least to discard unneeded information and metadata that are not even visible on the page, plus further standardization), 
we can instead implement these steps by designing a linearization of MusicXML and implementing conversion procedures. 
Aside from the pre- and post-processing requirements, this allows us to reformat the input data in a way that is much more amenable to sequence-to-sequence learning.

Similar encoding approaches have already been applied for monophonic music, most notably in the PrIMuS dataset \cite{calvo-zaragoza-etal-2018-end-to-end}. Recently, sequential encoding has been tried on homophonic \cite{AlfaroContreras2023} and pianoform music \cite{Rios-Vila-etal-2023-end-to-end}. Building on this work, we propose the LMX format: Linearized MusicXML.

\begin{figure*}[t]
    \centering
    \includegraphics[width=\hsize]{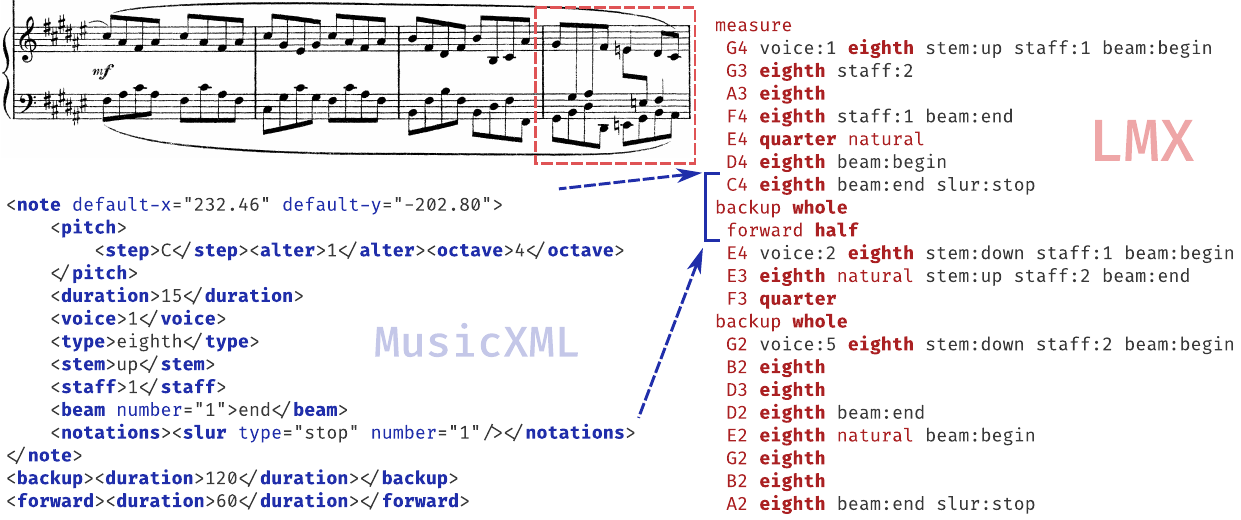}
    \caption{One measure -- 246 lines of MusicXML represented only by 96 tokens of Linearized MusicXML (formatting and indentation is present only for better readability).}
    \label{fig:lmx-xml-sample}
\end{figure*}

Thus, we propose the LMX format: Linearized MusicXML.
The initial idea is to take the XML tree, perform a depth-first walk over its nodes, and convert each element to a corresponding token. We leverage the fact that MusicXML defines the element order, which in turn defines the order of our sequence tokens. 
We also utilize existing element (and value) names and use them during the naming of our tokens.
Modifications are done to reduce verbosity (without loss of information). We encode some elements only via their values, for example, the note \verb`<type>` element is represented by tokens \verb`whole`, \verb`quarter`, \verb`16th`. We do not need to have an explicit \verb`type` token since it would convey no additional meaning. We only use the values as tokens in most other situations (\verb`G4` for pitch), often adding the type explicitly to aid LMX readability (\verb`voice:1`, \verb`tied:start`, \verb`clef:F2`).
We also reduce the number of tokens by only encoding state changes instead of absolutes for certain note properties. These are the voice number, staff number (staff within the grandstaff), and stem orientation. All of these are "forgotten" and re-emitted with each measure and with each voice change. We also omit the \verb`beam:continue` token, since it can be inferred.

MusicXML is designed not only for capturing notation visually but also for music replayability \cite{CalvoZaragoza-2020-understanding}. This means that visual information is often separated from audible information. A great example is the \verb`<tie>` and \verb`<tied>` elements. The first one specifies that the note's duration is extended to blend with the following note when played, whereas the second one states that there is a graphical tie present in the score. In this case, we only encode the visual \verb`<tied>` element in LMX and reconstruct both of them during de-linearization. Similarly, we do not encode the pitch \verb`<alter>` token, instead we encode \verb`<accidental>` and key signatures and later infer the alteration during decoding.

For replayability especially, MusicXML contains \verb`<duration>` elements, that encode time information in the number of \verb`<divisions>` (specified for the whole document). This duration can be reconstructed from the note's \verb`<type>`, duration dots, and \verb`<time-modification>` so we completely omit this information. The only problem is with \verb`<forward>` and \verb`<backup>` elements (note-like objects that only move the internal clock). These lack the \verb`<type>` element so the inference cannot be done here. Instead, we encode the duration as a combination of \verb`<type>` values, that together resolve to the same duration. This also unifies the representation of notes and rests, with these note-like elements and makes the \verb`<type>` token the root of any note-like object.

From this discussion, the design principles behind LMX can be summarized:
(1) minimize originality -- stick to MusicXML as much as possible;
(2) reduce excessive verbosity -- represent state changes, instead of state;
(3) focus on the visual aspect of music notation -- suppress semantics and ignore sound, layout, and metadata information.

Why MusicXML, and not some other format? A seemingly good linear representation is LilyPond, but its purpose is to typeset music, not to describe music that has already \emph{been} typeset. This makes it a programming language rather than a data format and it drives certain decisions, such as that one can force an accidental to appear by writing \verb`!`, but cannot represent an explicitly missing accidental. The Humdrum **kern is also a good option and it is used in the GradStaff dataset \cite{Rios-Vila-etal-2023-end-to-end}, but it cannot represent certain situations, such as a voice changing staves in the middle of a beamed group - a situation that should not be neglected in piano music. The remaining MEI and MusicXML are both mature, standardized, and well-known formats. We chose MusicXML because it is well supported by the open-source notation editor MuseScore and the same editor was chosen by Gotham et al. \cite{GothamJonas2022} for the creation of the OpenScore Lieder corpus, signifying its importance.

MuseScore also serves an important role in our setup: MusicXML canonicalization. While the standard defines a lot of properties, it leaves some to the user. These include within-chord note ordering, voice numbering (and ordering), or the specifics of linearization of multiple voices. This also includes hacks that get around MusicXML limitations (usually from converging voices on one note/chord) and various bugs and oddities of MuseScore (first voice rests cannot be deleted, only made invisible). It is important to state that we use MuseScore~3.6.2 and that a dedicated canonicalization module should be added in the future to get rid of the tight coupling with MuseScore.

\looseness-1
While MusicXML allows for arbitrary time-travel with the \verb`<forward>` and \verb`<backup>` elements, and for arbitrary note-voice assignment with the \verb`<voice>` element, it is more efficient for the format to represent voices one after each other within a single measure (though not the only option, **kern orders notes onset-wise, not voice-wise). MuseScore outputs exactly this variant, using the \verb`<backup>` command as a jump to the start of the measure and the next voice. It never uses it to jump smaller distances. If a voice starts or terminates inside a measure, \verb`<forward>` is added around the notes so that the voice takes up exactly one measure worth of duration. MuseScore also defines a maximum of 4 voices per staff and they are labeled 1--4 and 5--8 for two staves. We also adopt this approach.

One final interesting aspect is the encoding of tuplets. We encode the visual grouping of tuplets via \verb`tuplet:start` and \verb`tuplet:stop` and the duration change via an \verb`XinY` token derived from the \verb`<time-modification>` element. This token modifies the \verb`type` token so an eighth-note sextuplet has duration \verb`eighth`~\verb`6in4`. We do not cover nested tuplets. 
The LMX format supports 224 unique tokens.
Our current implementation discards dynamics markings, barline styles, pedal symbols, and other symbols with no effect on musical semantics; this accounts for about 4 \% of MusicXML nodes (Tab.~\ref{tbl:results_olimpic}).
The complete documentation of how LMX linearization works can be found in our GitHub repository.\footnote{\url{https://github.com/ufal/olimpic-icdar24}}

\section{Datasets}
\label{sec:datasets}

Existing OMR datasets fall roughly into two categories based on their purpose: object detection and end-to-end recognition. Object detection datasets contain very little pianoform music (only 3 pages in MUSCIMA++ \cite{MUSCIMAPP,CVC-MUSCIMA}), and the end-to-end datasets have focused mostly on monophonic or homophonic scores, such as PrIMuS \cite{calvo-zaragoza-etal-2018-end-to-end,camera_primus} and Alfaro-Contreras dataset \cite{AlfaroContreras2023}. The GrandStaff dataset is the first one targeting pianoform music \cite{Rios-Vila-etal-2023-end-to-end}. Another large set of manually encoded music that involves piano is the OpenScore Lieder corpus \cite{OpenScore.2018,GothamJonas2022}.

\subsubsection*{GrandStaff-LMX}

The GrandStaff dataset is a recently published, synthetic, and the first available pianoform dataset intended for end-to-end OMR \cite{Rios-Vila-etal-2023-end-to-end}. It is based on the KernScores corpus\footnote{\url{http://kern.ccarh.org/}} -- a collection of music scores in the Humdrum **kern format.
It contains 474 full-length scores by 6 composers, that were transposed to 3 additional key signatures and sliced up into 3--6 measure segments. These segments emulate individual systems\footnote{A system in music notation means one line of music, containing all the voices and instruments. It equals one grandstaff in this case.} of music on a page. One such segment represents one training sample for the end-to-end recognition model. The resulting dataset contains 53,882 data samples.
Each sample is accompanied by a synthetic JPG image of the music (rendered by Verovio \cite{Pugin_2014}) and its distorted variant. We refer to these distorted images as the Camera-GrandStaff dataset.

The authors of GrandStaff purposefully removed dynamics markings, slurs, lyrics, and non-graphic information **kern tokens. Each sample (system) also starts with clefs and key signature (as is usual in music notation), but also with time signature, which is not usually done in printed music. This may artificially help the model in the recognition of tuplets.

For the purpose of the experiments presented in this work, we convert the original GrandStaff encoding into MusicXML by the Music21 library and then into LMX (see Sec.~\ref{sec:lmx}). After linearization, the produced LMX files contain 133 unique tokens, omitting slurs, fermatas, tremolos, and many ornaments (staccato, arpeggios, accents). The resulting files are available for download at \url{http://hdl.handle.net/11234/1-5423}. 

\subsubsection*{OLiMPiC}
\label{sec:olc}
The OpenScore Lieder corpus \cite{OpenScore.2018,GothamJonas2022} is a collection of 19th-century German and French songs manually transcribed via MuseScore and made available in its MSCX format. We work with the corpus snapshot from Oct~30, 2023, which 
includes 1,356 scores (songs), coming from 253 sets\footnote{A~set is the extended work a song belongs to; e.g. a print edition.} by 107 composers, making it very diverse. Almost all scores have one voice part and an accompanying piano part. 

\looseness-1
We first used MuseScore~3.6.2 to convert the corpus to MusicXML and to generate PNG and SVG images.
We used the SVG output to detect the piano brace shape and match it with the corresponding stafflines. This let us slice the PNG files into individual systems, which would then be paired with the structured representations.
Not all scores could be processed in this way: some contain no piano part, some contain the brace symbol for non-piano parts, and others are problematic in many different unique ways. We had to skip 52, giving us 1,295 scores.\footnote{To ensure train-test set-independence, 9 more scores are ignored.}
We then extracted piano parts from the MusicXML and sliced them into pages and systems. We made sure each system starts with clefs and key signature (as it should). Finally, we used our linearizer to produce LMX annotations for each system. 
We release this processed subset as the OLiMPiC (\textbf{O}penScore \textbf{Li}eder \textbf{Li}nearized \textbf{M}usicXML \textbf{Pi}ano \textbf{C}orpus) dataset -- the synthetic variant. It contains 17,945 samples (music systems) with 182 unique LMX tokens.

The scores in OpenScore Lieder also contain a reference to the original IMSLP document they were transcribed from. Thanks to the strict transcription rules set for OpenScore, these IMSLP documents have an identical score layout as the transcriptions (measures per system, systems per page). We manually annotated system bounding boxes for 200 scores in their original IMSLP documents to acquire real-world scanned images (dev and test set). 

We extracted the PNG images from IMSLP PDFs using the \verb`pdfimages` tool\footnote{Available from \url{https://www.xpdfreader.com/}}. This sidestepped any rasterization losses but caused annotation issues because some images were extracted B/W inverted, some were rotated, and some were split to multiple layers making them unusable as-is. 
Also, sometimes MuseScore wraps a system too early, creating an additional system and de-synchronizing the layout for the score. This is not an issue in synthetic images, but here we decided to skip these cases instead of painfully implementing a fix. This made us skip 60 scores before we were able to annotate the desired 200 scores. Given the reasons for skipping, we believe this did not introduce any bias to the test set. We release this dataset as OLiMPiC -- the scanned variant.

Both the scanned and synthetic OLiMPiC variants come with the same train/dev/test splits. These splits are defined by score IDs and the test split is also set-independent, meaning if a score appears in the test set, no other score from the same set is allowed to appear in the train/dev sets. The sizes of training, dev, and test sets are shown in Tab.~\ref{tbl:olimpic_stats}.
The complete dataset is available for download at \url{http://hdl.handle.net/11234/1-5419} under the CC BY-SA license.

\begin{figure*}[t]
    \centering
    \includegraphics[width=\hsize]{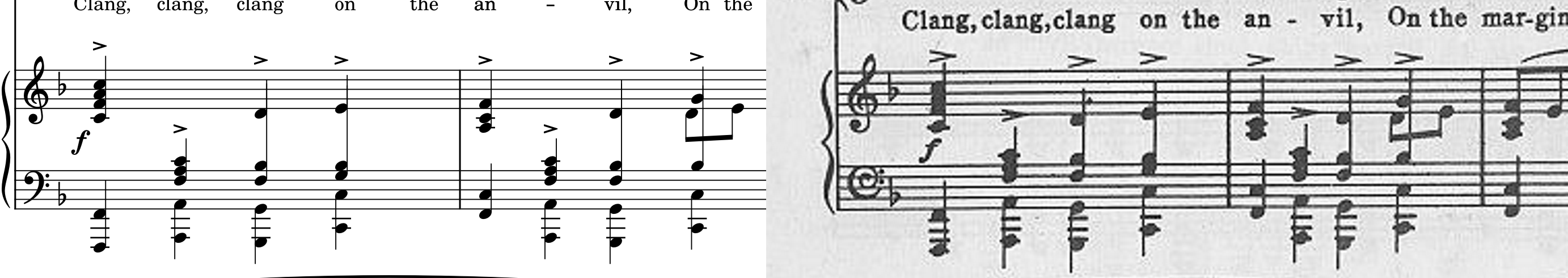}
    \caption{Comparison of a synthetic and scanned sample. Notice the different bass clef style and measure width.}
    \label{fig:olimpic-sample}
\end{figure*}

\begin{table}[bt]
  \caption{The OLiMPiC dataset statistics.}
  \label{tbl:olimpic_stats}
  \def\0{\hphantom{0}}
  \centering
  \begin{tabular}{l@{\kern1em}c@{\kern1em}c@{\kern1em}c@{\kern1em}c@{\kern1em}c@{\kern1em}c}
      \toprule
      \textbf{Partition} & \textbf{Synthetic} & \textbf{Scanned} & \textbf{\#Sets} & \textbf{\#Scores} & \textbf{\#Samples} & \textbf{\#Tokens} \\
      \midrule
      Train & \YES & \NO  & 206 & 1,095 & 15,014 & 4,107,597 \\
      Dev   & \YES & \YES & \071 & \0~100 & \01,438 & \0~378,119 \\
      Test  & \YES & \YES & \033 & \0~100 & \01,493 & \0~405,104 \\
      \bottomrule
  \end{tabular}
\end{table}

\section{Evaluation}
\label{sec:evaluation}

Evaluation of OMR is an open problem with few solutions in sight, much less in practice \cite{Bellini2007,Byrd2010,Byrd2015,Hajicjr.2016,Hajicjr.2018b,CalvoZaragoza-2020-understanding}.
There is no one overall best way to evaluate OMR systems because user needs differ significantly among use cases.
Our focus in this work is on effort-to-correct (also known as recognition gain \cite{Bellini2007}):
How much work would it be for a user to post-process the output to match the desired music?
(Again, this can be approximated just very roughly.)

For sequence-to-sequence models, the go-to class of evaluation metrics is the Symbol Error Rate (SER)
which counts the proportion of correctly predicted symbols, and the more stringent
Line Error Rate (LER) that counts the proportion of error-free lines.
While this is a natural choice, especially during development, when an automated metric is necessary,
interpreting the SER numbers is not straightforward. First, symbols in an encoding might have vastly different
importance to the result (even before a user is considered) \cite{Bellini2007}: 
some may influence the semantics of just one note, 
others may influence multiple (clef and key signature errors, notoriously, or tuples), 
others may be negligible (such as the presence or absence of articulation marks). 
While the specific weights assigned
to error classes should take specific use cases into account, which can hardly be done in the course of basic research,
we can at least broadly say that errors that influence the musical semantics of the recognition output -- pitches, durations,
and ordering of notes (or the absence of one, or the presence of a spurious note) -- should perhaps have more weight.
Second, the choice of linearization introduces artifacts, such as "advance" characters 
after every note \cite{AlfaroConterras2019} or dots for empty positions 
on unused spines in **kern \cite{Rios-Vila-etal-2023-end-to-end}).
The presence of such artifacts further complicates the comparison between systems that use different linearizations.

A practical comparison of OMR systems should compare ``apples to apples'' \cite{Bellini2007,Byrd2015}.
Despite the advantages of LMX for linearization, it is unavoidable that other design criteria will
lead developers to use different encodings, and anyway, ideally the comparison of systems should not depend
on the choice of linearization (which is merely a necessity dictated only by the currently best-performing class
of models), so this kind of representation is in principle not suitable.
Such an ``apples to apples'' comparison can be done at the level of correctly recovering the finite information 
visually encoded on the page \cite{Hajicjr.2018b}, as exemplified mostly by correct symbol counting \cite{Droettboom2004,Bellini2007},
but this approach cannot be used with end-to-end systems that do \textit{not} recover 
explicit information about the placement of individual
notation symbols. 

The other option that avoids polluting evaluation metrics with artifacts of the specific
methods used, and possibly more informative for a user, is to compare directly the true endpoint
of the OMR process -- the encoding of the output in a broadly usable format \cite{CalvoZaragoza-2020-understanding}.
MusicXML is a natural choice for this purpose. Evaluating OMR by comparing MusicXML representations has in fact been proposed 
for this purpose \cite{Knopke2007,Padilla2014}.\footnote{For applications focused on the ``musical semantics'' of notes only, without regard for what elements of music notation were used to encode them, the natural endpoint would be MIDI, such as in \cite{Hajicjr.2018a}.} So far, however, there is little consensus on \textit{how} to compare two MusiXML files
(Padilla et~al. write that they "align the OMR output to the ground truth",
with no further details provided \cite{Padilla2014}).

XML files are organized as trees, so Tree Edit Distance (TED) would be a natural choice. Polynomial-time algorithms are known, 
esp. the Zhang-Shasha algorithm that runs in $O(m^2n^2)$ time and only requires $O(mn)$ memory complexity \cite{Zhang1989} 
which also has a Python implementation \texttt{zss}. 
Zhang-Shasha relies on the ordering of child nodes, but fortunately, MusicXML does have an ordering of child nodes defined,
and the remaining ambiguity is handled by MusicXML canonization described in
Haji\v{c} jr.~et~al. \cite{Hajicjr.2016} proposes TEDn, 
a modified TED to estimate replacement costs for differences in musical semantics on notes specifically 
(which naive TED would over-estimate because of how MusicXML decomposes this information into nodes),
and, importantly, provides evidence that TEDn correlates with human editors' expectations 
of how much effort it would take to modify one file to fit the other using a WYSIWYG editor like MuseScore
better than the (few) alternatives.

Therefore, we created a new TEDn implementation and use it as an evaluation metric in this work. 
One disadvantage of TEDn is that its behavior with respect to SER is not understood,
as TEDn has not yet in fact been used to evaluate experiments (because the end-to-end models have not
yet been producing MusicXML outputs). 
Therefore, we report SER as well, which also allows us to directly compare to previous work, 
and also to have a more direct comparison of the GrandStaff and OLiMPiC datasets.

\section{Experiments}

\looseness-1
In the neural network era, optical character recognition (OCR) has been commonly approached by using the convolutional recurrent neural network (CRNN) model~\cite{shi-etal-2017-an-end-to-end}. In this model, an image with a line of text is first passed through several convolutional layers, then processed with bidirectional recurrent neural networks~\cite{graves-schmidhuber-2005-framewise}, most commonly LSTMS~\cite{hochreiter-schmidhuber-1997-long,gers-etal-1999-continual}, and then a prediction is performed using the CTC layer~\cite{graves-etal-2006-connectionist}. The CTC layer enables efficient training and inference when an output sequence (characters in case of OCR) should be generated from an input sequence (representation of fixed-with columns on the input image) without an explicit alignment; however, it requires the order of elements to be the same in both input and output sequences.

\looseness-1
The CRNN model can be extended to OMR, but the requirement of the same ordering in the input and output sequences limits it only to homophonic scores~\cite{calvo-zaragoza-etal-2018-end-to-end,calvo-zaragoza-etal-2018-handwritten,AlfaroConterras2019}. To approach optical music recognition of polyphonic music, we exchange the CTC layer of the CRNN model with a sequence-to-sequence architecture~\cite{sutskever-etal-2014-sequence,cho-etal-2014-learning}, namely an LSTM decoder with Bahdanou attention~\cite{bahdanau-etal-2015-neural}.

\begin{figure*}[t]
    \centering
    \includegraphics[width=\hsize]{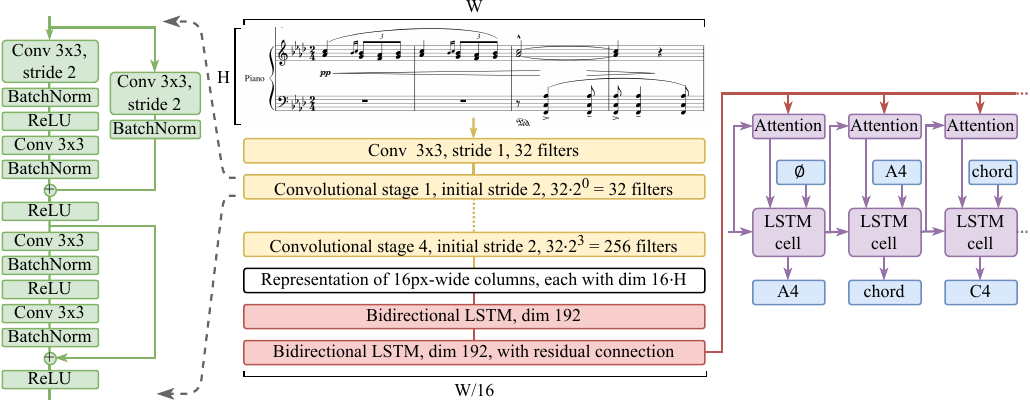}
    \caption{Architecture of our model.}
    \label{fig:architecture}
\end{figure*}

The architecture of our model dubbed Zeus is detailed in Fig.~\ref{fig:architecture}. The input fixed-height image is first passed through a single $3\times3$ convolution, and then through four convolutional stages. In each stage, two ResNet-like blocks~\cite{he-etal-2015-deep} with batch normalization~\cite{ioffe-szegedy-2015-batch} are employed, with first convolution in a stage having a stride 2 and since stage 2 doubling the number of filters. Afterward, the values in a single image column are concatenated, obtaining representations of fixed-width columns in the input image. These representations are contextualized by two layers of bidirectional LSTM~\cite{graves-schmidhuber-2005-framewise}, the second with a residual connection, and finally used as input to an LSTM decoder with Bahdanau attention~\cite{bahdanau-etal-2015-neural}.

\looseness-1
We deliberately do not use the Transformer architecture~\cite{vaswani-etal-2017-attention}, neither as the image encoder nor the sequence decoder. While it is capable of delivering unrivaled performance, it requires a substantial amount of data (even the data-efficient masked autoencoders~\cite{he-etal-2022-masked} employ a million images), and we surmise both the sequence encoder and decoder benefit from the inductive locality bias of LSTMs. In Section~\ref{sec:results}, we validate our approach by showing that our model delivers a 50\% relative error reduction compared to existing RNN- and Transformer-based models.

We train the model on a single 40GB A100 GPU using the Adam optimizer~\cite{kingma-ba-2015-adam} and a learning rate of 1e-3 with a cosine-decay~\cite{loshchilov-hutter-2017-sgdr} for 500 epochs with a batch size of 64. The input image is rescaled to height 192, the dimensionality of the LSTM cells is set to 192, and we use dropout of 0.2 before and after the bidirectional LSTM layers.

\subsubsection*{Augmentations} 

\begin{figure*}[t]
    \centering
    \includegraphics[width=.2\hsize]{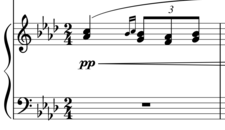}%
    \includegraphics[width=.2\hsize]{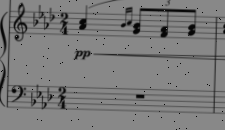}%
    \includegraphics[width=.2\hsize]{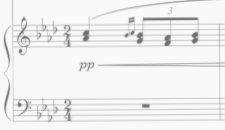}%
    \includegraphics[width=.2\hsize]{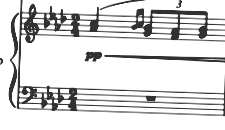}%
    \includegraphics[width=.2\hsize]{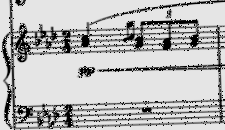}
    \caption{An exemplary part of a system and its four random augmentations.}
    \label{fig:augmentations}
\end{figure*}

Given that the training data is synthetic and our goal is to process scanned images, we optionally apply augmentation operations to the synthetic training data. For every image, we consider the following operations in the given order and apply each with 50\% chance and randomly chosen magnitude:
\begin{citemize}
    \item horizontal shift by at most 8 pixels,
    \item rotation by at most 1 degree,
    \item vertical shift by at most 4 pixels,
    \item dilatation/erosion in a random direction on an ellipse with x semi-axis 1 and y semi-axis 0.5,
    \item for a random probability of up to 20\%, negate pixels whose value and value of their 8 neighbors are not uniformly white or uniformly black,
    \item for a random probability of up to 1\%, negate every pixel,
    \item adjust contrast by a factor with random base-2 logarithm  in $[-1, 1]$ range,
    \item adjust brightness by a random factor in $[-0.5, 0.2]$ range.
\end{citemize}
Four random augmentations of a part of a system are displayed in Fig.~\ref{fig:augmentations}. For more details, see the source code of the implementation.

\section{Results}
\label{sec:results}

\subsubsection*{GrandStaff}

We first show that our model surpasses existing models for optical pianoform music recognition, giving credibility to the later results on the OLiMPiC dataset.
We compare our model to the three architectures proposed in Rios-Vila et al.~\cite{Rios-Vila-etal-2023-end-to-end} on the GrandStaff and Camera Grandstaff datasets. The evaluation is performed using Character Error Rate (CER), Symbol Error Rate (SER), and Line Error Rate (LER).\footnote{\label{fnt:cer_bug}The Character Error Rate originally computed in \cite{Rios-Vila-etal-2023-end-to-end} contained a bug that has since been fixed in \url{https://github.com/multiscore/e2e-pianoform}. We report this CER$_\mathrm{bug}$ to compare directly to \cite{Rios-Vila-etal-2023-end-to-end}, but we also report the correct CER for future comparison.}

The results are presented in Table~\ref{tbl:results_grandstaff_original}. When trained on GrandStaff and also on Camera GrandStaff, our approach reduces the errors by at least 50\% relative compared to all other models, including the Transformer one.

\begin{table}[t]
  \caption{Evaluation of our model on the GrandStaff dataset~\cite{Rios-Vila-etal-2023-end-to-end}.\textsuperscript{\ref{fnt:cer_bug}}}
  \label{tbl:results_grandstaff_original}
  \def\0{\hphantom{0}}
  \catcode`! = 13\def!{\bfseries}
  \centering
  \setlength{\tabcolsep}{0.6ex}
  \begin{tabular}{lcccc@{\kern.9em}cccc@{\kern.5em}}
      \toprule
      \multirow{2}{*}{\bfseries Model} & \multicolumn{4}{c}{\bfseries \kern-1em GrandStaff [\%]} & \multicolumn{4}{c}{\bfseries Camera-GrandStaff [\%]} \\
      & \llap{CE}R$_\mathrm{bug}$ & CER & SER & LER & CER$_\mathrm{bug}$ & CER & SER & LER \\
      \midrule
      Encoder-only CNN~\cite{Rios-Vila-etal-2023-end-to-end} & \06.4 & --- & 11.3 & 29.8 & 11.9 & --- & 22.5 & 58.3 \\      
      CNN, RNN decoder~\cite{Rios-Vila-etal-2023-end-to-end} & \05.0 & --- & \07.3 & 23.2 & \07.2 & --- & \09.9 & 29.5 \\
      CNN, Transformer decoder~\cite{Rios-Vila-etal-2023-end-to-end} & \03.9 & --- & \05.8 & 16.3 & \04.6 & --- & \06.5 & 17.5 \\
      \midrule
      Zeus & !\01.6\rlap{8} & !2.3\rlap{0} & !\02.7\rlap{7} & !\08.1\rlap{9} & !\01.9\rlap{1} & !2.5\rlap{4} & !\03.0\rlap{3} & !\08.4\rlap{9} \\
      \bottomrule
  \end{tabular}
\end{table}

\begin{table}[t]
  \caption{The results on the OLiMPiC and GrandStaff-LMX datasets.}
  \label{tbl:results_olimpic}
  \def\0{\hphantom{0}}
  \centering
  \begin{tabular}{lc@{\kern1em}c@{\kern.5em}c@{\kern1.75em}c@{\kern.9em}c}
    \toprule
    \textbf{Dataset} & \textbf{Augmented} & \makecell[c]{\textbf{SER}\\\textbf{full [\%]}} & \makecell[c]{\textbf{SER w\kern-1pt/\kern-1pt\rlap{o}}\\\textbf{tuplets [\%]}} & \makecell[c]{\textbf{TEDn}\\\textbf{full [\%]}} & \makecell[c]{\textbf{TEDn}\\\textbf{lmx [\%]}} \\
    \midrule
    OLiMPiC Synthetic & \NO  & 11.29  & \09.89  & 13.74  & \09.89 \\
    OLiMPiC Synthetic & \YES & 12.04  & 10.48  & 14.41  & 10.57 \\
    %\midrule
    OLiMPiC Scanned   & \NO  & 59.90  & 58.11  & 44.41  & 42.45 \\
    OLiMPiC Scanned   & \YES & 17.72  & 16.11  & 18.40  & 14.85 \\
    \midrule
    GrandStaff-LMX    & \NO  & \01.78 & \01.70 & \01.60 & \01.56 \\
    Camera GrandStaff-LM\rlap{X}     & \NO  & \01.99 & \01.92 & \01.77 & \01.73 \\
    \bottomrule
  \end{tabular}
\end{table}

\subsubsection*{OLiMPiC}

The performance of our model on the OLiMPiC dataset is quantified in Table~\ref{tbl:results_olimpic}. We train two models -- without the training data augmentations and with them and evaluate on both the synthetic and scanned test sets. We report SER on the full Linearized MusicXML, and also the TEDn metric using both the full MusicXML and only the subset captured by Linearized MusicXML. Out of these alternatives, only the full TEDn metric is independent on the encoding selected and capable of comparing dissimilar models.

Considering first the model without augmentations, it achieves 11.3\% SER on the synthetic dataset. The TEDn metric evaluated on the full MusicXML is 13.7\% and decreases by nearly 4 percent points when considering only the subset captured by Linearized MusicXML. Unsurprisingly, when the model is applied to the scanned images, it performs poorly with 44.4\% full TEDn.

The model with augmentations performs slightly worse on the synthetic dataset -- by less than a percent point absolute. However, its performance on the scanned images improves considerably to 18.4\% full TEDn (one-third more errors compared to the synthetic dataset) and 14.85\% Linearized-MusicXML-specific TEDn (one-half more errors). This setting, in our view, starts to provide meaningful numbers in measuring OMR performance overall. Given the inherent limitations of manually annotating real-world images, a user is likely to bring out-of-domain images. The scanned test images simulate this expected out-of-domain nature of production scenarios, at least for IMSLP-style repositories of printed music PDFs, because the OLiMPiC test set comprises flatbed scans with little to no unevenness in lighting and 3D deformation and thus does not provide yet a good model for images taken with phones.

\begin{figure*}[t]
    \centering
    \begin{minipage}{.49\hsize}
    \includegraphics[width=\hsize,trim={120pt 150pt 160pt 120pt},clip]{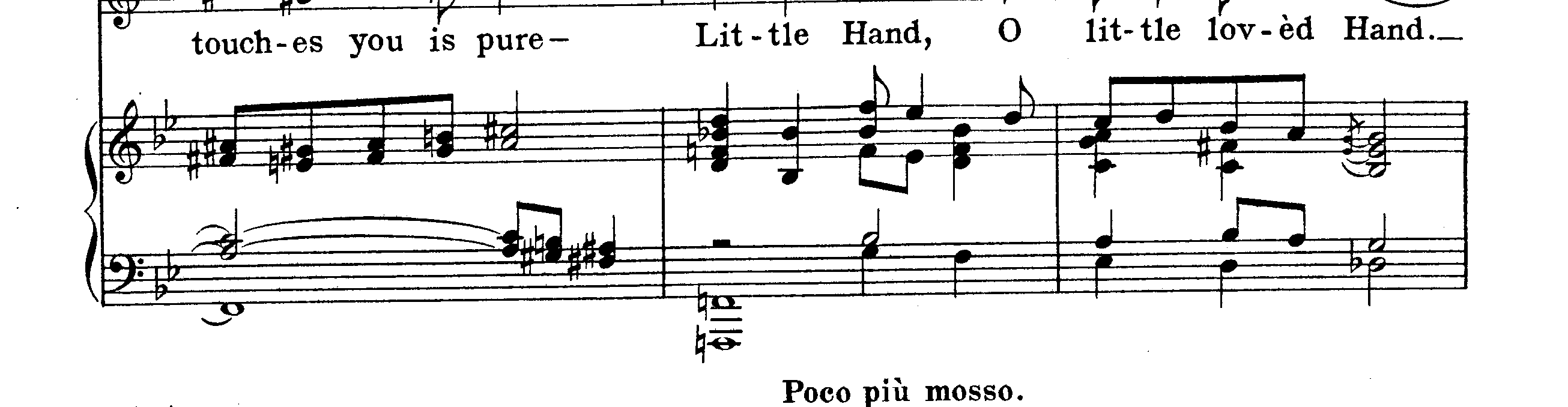}\vspace{5pt}
    \includegraphics[width=\hsize]{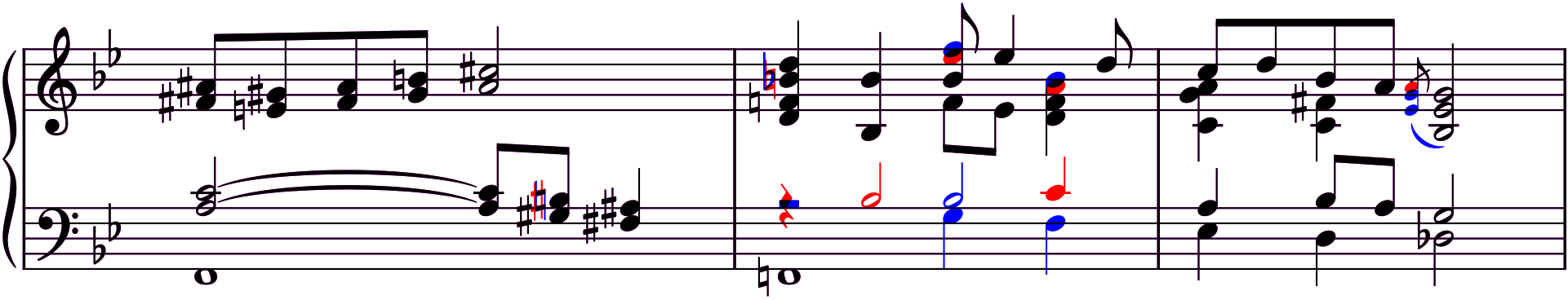}\vspace{5pt}
    \end{minipage}\kern.02\hsize%
    \begin{minipage}{.49\hsize}
    \includegraphics[width=\hsize,trim={10pt 55pt 200pt 140pt},clip]{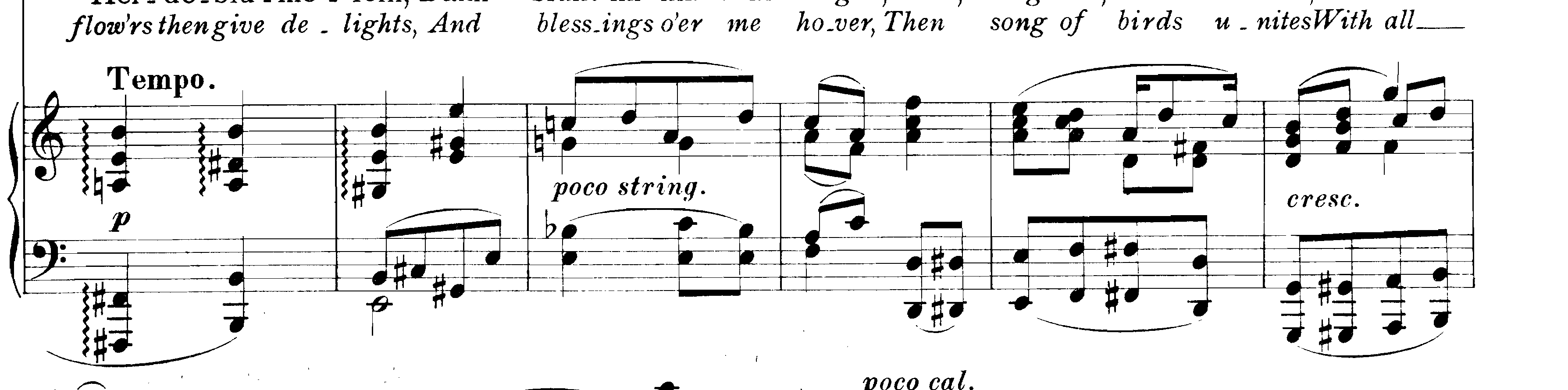}
    \includegraphics[width=\hsize]{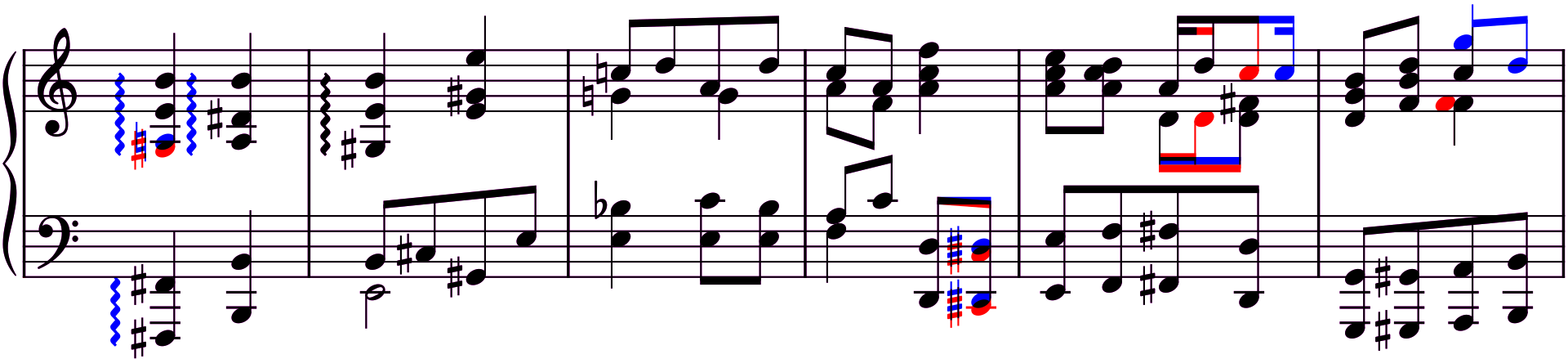}
    \end{minipage}
    \caption{A median-error recognition result on OLiMPiC scanned. Blue is ground truth.}
    \label{fig:recognition-result}
\end{figure*}

In Fig.~\ref{fig:recognition-result}, we show two example systems from the OLiMPiC scanned dataset and visualization of their recognition. We chose systems containing the median number of recognition errors.

\subsubsection*{GrandStaff-LMX} Table~\ref{tbl:results_olimpic} includes also the results on the (Camera) GrandStaff-LMX datasets. Both the SER and TEDn metrics on these datasets are less than one-fifth compared to the OLiMPiC dataset, supporting our claim that the music itself in OLiMPiC is a significantly harder pianoform OMR challenge. At the same time, being based on the real-world nature of the OpenScore Lieder Corpus, we believe the measurements on OLiMPiC to be a more accurate reflection of how state-of-the-art end-to-end OMR systems actually perform from a hypothetical user's perspective.    

\section{Discussion and Conclusions}

Our work enables applying state-of-the-art sequence-to-sequence models to process
pianoform music and output MusicXML, with only MusicXML representations of training data required.
Thus, it is now possible to directly develop OMR models for this practical,
broadly supported format for representing music notation. Additionally, all sheet music already available
in MusicXML (for instance via the MuseScore community) has been ``unlocked'' for OMR training.
Because of limitations of the existing GrandStaff dataset, chief among them being rather ``easy''
(as evidenced in Tab.~\ref{tbl:results_olimpic}), we derived the OLiMPiC dataset from
the OpenScore Lieder corpus, which can serve as a sufficiently difficult benchmark 
for comparing further pianoform OMR.
While our datasets are pianoform, we worked on this music as it is the most complex, and thus
the tools for handling it are as general as necessary to process other kinds
of CWMN. Note also that while our experiments are done on individual systems, 
not entire pages, that is a property only of the experimental setup -- nothing
in the (de)linearization or evaluation procedures requires splitting the page into systems.

The TEDn evaluation metric then allows for an apples-to-apples comparison 
directly on MusicXML files, regardless of the linearization or other intermediate
representations used within competing systems.

We therefore believe we are now significantly closer to establishing
an objective methodology for comparing different OMR systems
-- again, directly in a broadly adopted interchange format. 
While more needs to be done to understand the interactions 
between the ZSS algorithm and various weighing schemes for elements of MusicXML, 
there are at least results that show it also correlates with human editors' preferences
\cite{Hajicjr.2016}, and thus represents the best available metric.

Finally, our experiments achieve state-of-the-art results on piano music, even with less
resource-intensive attention model than the Transformer architecture.
They demonstrate that the LMX-based pipeline introduces no new risk for training 
the sequence-to-sequence models at the core of improvements in OMR performance,
while presenting significant advantages in practicality. Also, these results
can serve as a baseline for further development and improvements in OMR models -- perhaps challenging,
but hopefully not be too hard to overtake. 
If there is a model next month that performs better on the OLiMPiC dataset
using LMX and reporting on TEDn, we will consider this work successful than if we
retain ``top score'' for longer.

\subsubsection*{Limitations and Future work}

One serious limitation of our work is that we rely on MuseScore 3.6.2 for MusicXML canonization, 
and thus we have a critical external -- albeit open-source -- dependency. While not an immediate
issue, it will take significant development effort to implement MusicXML canonization ourselves,
so that LMX becomes a truly standalone, transparent toolchain.

For the recognition model, tuplets remain the most serious issue, with the \texttt{2in3}
symbol accounting for about 1.5\% of SER. This is due to ``implicit'' triplets:
note groups beamed in groups of 3 that should be played as triplets (obviously to human players)
but are not explicitly marked as such.

Within the LMX encoding, aside from the roughly 4\% of TEDn performance due to LMX not covering
certain symbols (see Tab.~\ref{tbl:results_olimpic}), sets of slurs that reach across multiple measures in 
parallel are not delinearized in the correct order; we expect that with broader adoption and new datasets,
such bugs and edge cases will be discovered. The open-source licensing of our code
fortunately allows addressing such limitations as they are encountered by the community.

Finally, we have been glossing over the distinction between OMR for reprintability and replayability,
and the different purposes for which OMR can be used \cite{CalvoZaragoza-2020-understanding}.
Each of these indeed imposes different
evaluation criteria: for instance, focusing on retrieval based on melodies alone does not much care
for the correctness of articulation marks, how notes are assigned to voices, 
or whether other notes than the melody are recognized at all. However, the MusicXML format
that we selected with practicality in mind requires the OMR system to recover both the musical semantics,
and (most of) the elements of music notation used to encode this music. In any case, adaptations that 
reduce the vocabulary of LMX for users with such more specific needs are straightforward to implement.

The contributions that we present here should also finally make it possible
to move towards one of the major goals of the OMR community, as stated by Calvo-Zaragoza, Haji\v{c} jr.
and Pacha after a seminal community meeting at GREC/ICDAR 2017 \cite{Calvo-Zaragoza2018d}:
greater interoperability. Perhaps the long-sought goal of creating an OMR benchmark \cite{Byrd2015}
that communicates meaningful answers to the question: "Does OMR work?" is now within reach.

\bibliographystyle{splncs04}
\bibliography{bibliography}

\end{document}